\documentclass{article} %

\newif\ifarxiv
\arxivtrue

\newif\ifperfect
\perfectfalse %

\usepackage{times}
\usepackage{latexsym}
\usepackage[T1]{fontenc}
\usepackage[utf8]{inputenc}
\usepackage{microtype}

\ifarxiv
    \usepackage{acl}
\else
    \usepackage[review]{acl}
\fi

\usepackage{amsmath,amsfonts,bm}

\def\eqref#1{equation~\ref{#1}}

\def\1{\bm{1}}

\DeclareMathAlphabet{\mathsfit}{\encodingdefault}{\sfdefault}{m}{sl}
\SetMathAlphabet{\mathsfit}{bold}{\encodingdefault}{\sfdefault}{bx}{n}

\newcommand{\independent}{\perp\!\!\!\perp}

\usepackage{hyperref}
\usepackage{url}
\usepackage{mathtools}
\usepackage{colortbl}
\usepackage{sec/customized_package}

\newcommand{\ourmodel}{{\modelfont{TextMatch}}\xspace}
\newcommand{\ourname}{\textsc{CausalCite}\xspace}

\newcommand{\pciavg}{ACI\xspace}

\graphicspath{ {./fig/} }
\usepackage{subcaption}
\usepackage{wrapfig, lipsum, booktabs}
\newtheorem{assumption}{Assumption}

\begin{document}
 
\title{\textbf{\ourname}: A Causal Formulation of Paper Citations}

\author{Ishan Kumar\textsuperscript{1}\thanks{{ } Equal contribution.
} { } { } Zhijing Jin\textsuperscript{1,2}\samethanks { } { } Ehsan Mokhtarian\textsuperscript{3} { }
\\
\textbf{
Siyuan Guo\textsuperscript{1,4} { } Yuen Chen\textsuperscript{1} { } 
Mrinmaya Sachan\textsuperscript{2} { } Bernhard Schölkopf\textsuperscript{1}} \\
\textsuperscript{1}Max Planck Institute for Intelligent Systems, Tübingen, Germany  \\
\textsuperscript{2}ETH Zürich \quad
\textsuperscript{3}EPFL \quad
\textsuperscript{4}University of Cambridge 
\\
\texttt{ishankumaragrawal@gmail.com} \quad \texttt{jinzhi@ethz.ch}
\\}

\maketitle

\begin{abstract}
Citation count of a paper is a commonly used proxy for evaluating the significance of a paper in the scientific community. Yet citation measures are widely criticized for failing to accurately reflect the true impact of a paper.
Thus, we propose \ourname, a new way to measure the significance of a paper by assessing the causal impact of the paper on its follow-up papers.
\ourname\ is based on a novel causal inference method, \textit{\ourmodel}, which adapts the traditional matching framework to high-dimensional text embeddings.  \ourmodel\ encodes each paper using text embeddings from large language models (LLMs), extracts similar samples by cosine similarity, and synthesizes a counterfactual sample as the weighted average of similar papers according to their similarity values. 
We demonstrate the effectiveness of \ourname\ on various criteria, such as high correlation with paper impact as reported by scientific experts on a previous dataset of 1K papers, (test-of-time) awards for past papers, and its stability across various subfields of AI. We also provide a set of findings that can serve as suggested ways for future researchers to use our metric for a better understanding of the quality of a paper.\footnote{\ifarxiv
Our code is available at \url{https://github.com/causalNLP/causal-cite}.
\fi}
\end{abstract}

\section{Introduction}
Recent years have seen explosive growth in the number of scientific publications, making it increasingly challenging for scientists to navigate the vast landscape of scientific literature.
Therefore, identifying a good paper has become a crucial challenge for the scientific community, not only for technical research purposes, but also for making decisions, such as funding allocation \citep{carlsson2009allocation}, research evaluation \citep{moed2006citation}, recruitment \citep{holden2005bibliometrics}, and university ranking and evaluation \citep{piro2016how}. 

\begin{figure}[t]
    \centering
    \includegraphics[width=\linewidth]{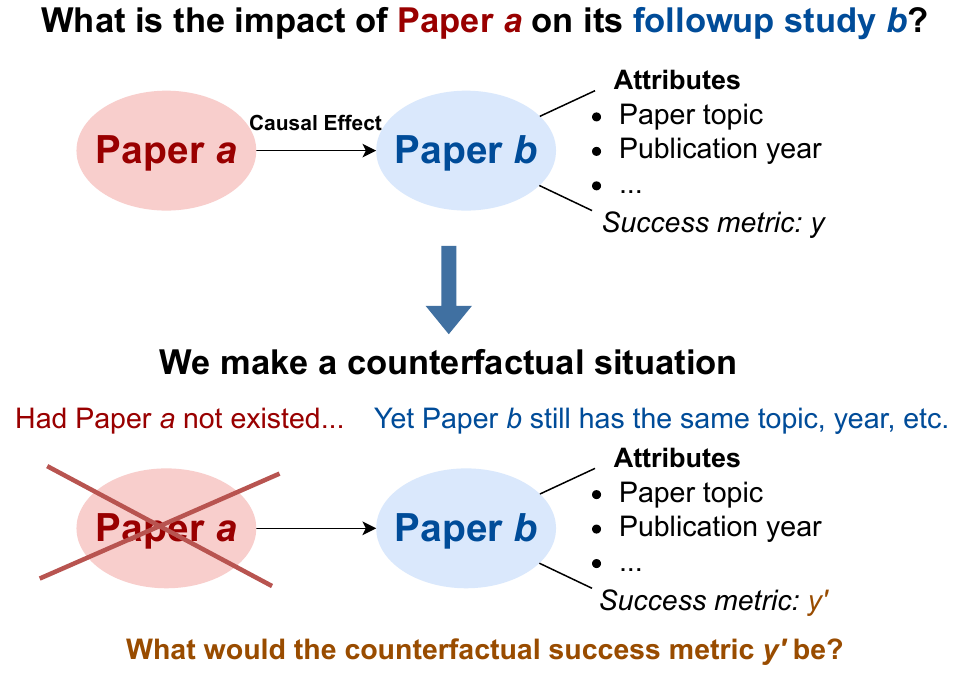}
    \caption{An overview of our research question.}
    \label{fig:intro}
\end{figure}

A traditional approach to recognize paper quality is peer review, a mechanism that requires large efforts, and yet has inherent randomness and flaws \citep{cortes2021inconsistency, rogers-etal-2023-report, shah2021survey, prechelt2018community, resnik2008perceptions}.
Moreover, the number of papers after peer review is still overwhelmingly large for researchers to read, leaving the challenge of identifying truly impactful research unaddressed.
Another commonly used metric is citations. However, this metric faces criticism for biases, such as a preference for survey, toolkit, and dataset papers \citep{zhu, semanticHICA}.
Together with altmetrics \citep{altMetrics}, which incorporates social media attention to a paper, both metrics also bias towards papers from major publishing countries \citep{rungta-etal-2022-geographic,gomez2022leading}, with extensive publicity and promotion, and authored by established figures.

To provide a more equitable assessment of paper quality, we employ the causal inference framework \citep{hernan2010causal} to quantify a paper's impact by how much of the academic success in the follow-up papers should be \textit{causally attributed} to this paper.
We introduce \ourname, an enhanced citation based metric that poses the following \emph{counterfactual} question (also shown in \cref{fig:intro}): ``\textit{had this paper never been published, what would have happened to its
follow-up studies?}''
To compute the causal attribution of each follow-up paper, we contrast its citations (the treatment group) with citations of papers that address a similar topic, but are not built on the paper of interest (the control group).

Traditionally, this problem is  solved by using the matching method \cite{rosenbaum1983central} in causal inference, which discretizes the value of the confounder variable, and compares the treatment and control groups with regard to each discretized value of the confounder variable. However, this approach does not apply when the confounder variable is
high-dimensional, e.g., text data, such as the content of the paper. %
Thus, we improve the matching method to adapt for textual confounders, by marrying recent
advancement of large language models (LLMs) with traditional causal inference.
Specifically, we propose \ourmodel, which uses LLMs to encode an academic paper as a high-dimensional text embedding to represent the confounders, and then, instead of iterating over discretized values of the confounder, we match each paper in the treatment group with papers from the control group with high cosine similarity by the text embeddings.

\ourmodel makes contributions in three different aspects: (1) it relaxes the previous constraint that the confounder variable should be binned into a limited set of intervals, and makes the matching method applicable for high-dimensional continuous variable type for the confounder; (2) since there are millions of papers, we enable efficient matching via a matching-and-reranking approach, %
first using information retrieval (IR) \citep{cosineSim} to extract a small set of candidates, and then applying semantic textual similarity (STS) \citep{majumder2016semantic,chandrasekaran2022evolution} for fine-grained reranking; and (3) we enable a more stable causal effect estimation by
leveraging all the close matches to synthesize the \textit{counterfactual citation score} by a weighted average according to the similarity scores of the matched papers. 

\ourname quantifies scientific impact via a causal lens, offering an alternative understanding of a paper's impact within the academic community.
To test its effectiveness, we conduct extensive experiments using the Semantic Scholar corpus \citep{lo-etal-2020-s2orc,Kinney2023TheSS}, comprising of $206$M papers and $2.4$B citation links.
We empirically validate \ourname by showing higher predictive accuracy of paper impact (as judged by scientific experts on a past dataset of 1K papers \citep{zhu}) compared to citations and other previous impact assessment metrics.
We further show a stronger correlation of the metric with the test-of-time (ToT) paper awards.
We find that, unlike citation counts, our metric exhibits a greater balance across various research domains in AI, e.g., general AI, NLP, and computer vision (CV). While citation numbers for papers in these domains vary significantly -- for example, while an average CV paper has many more citations than an average NLP paper, \ourname scores papers across AI sub-fields more similarly.

After demonstrating the desirable properties of our metric, we also present several case studies of its applications.
Our findings reveal that the quality of conference best papers is noisier on average than that of ToT papers (\cref{sec:best}). %
We then showcase and present \ourname for several well-known papers (\cref{sec:famous_papers}) and utilize \ourname to identify high-quality papers that are less recognized by citation counts (\cref{sec:outlier}).

In conclusion, our contributions are as follows:
\begin{enumerate}%
    \item We introduce \ourname, a counterfactual causal effect-based formulation for paper citations.
    \item We develop \ourmodel, a new method that leverages LLMs and causal inference to estimate the counterfactual causal effect of a paper.
    \item We conduct comprehensive analyses, including various performance evaluations and present new findings using our metric.
\end{enumerate}

\section{Problem Formulation}
Our problem formulation involves a citation graph and a causal graph. We use lowercase letters for specific papers and uppercase for an arbitrary paper treated as a random variable.

\paragraph{Citation Graph}
In the citation graph $\mathbb{G} \coloneqq (\mathbb{P}, \mathbb{L})$, $\mathbb{P}$ is a set of papers, and each edge $\ell_{i,j} \in \mathbb{L}$ indicates that an earlier paper ${p}_i$ influences (i.e., is cited by) a follow-up paper ${p}_j$.
To obtain the citation graph, we use the Semantic Scholar Academic Graph dataset \citep{Kinney2023TheSS} with 206M papers and 2.4B citation edges.
\ifperfect \red{We focus on non-trivial influence, limiting citation links to those identified as highly influential by Semantic Scholar \citep{ValenzuelaEscarcega2015IdentifyingMC}.}\fi

\begin{figure}[ht]
    \centering
    \includegraphics[width=\linewidth]{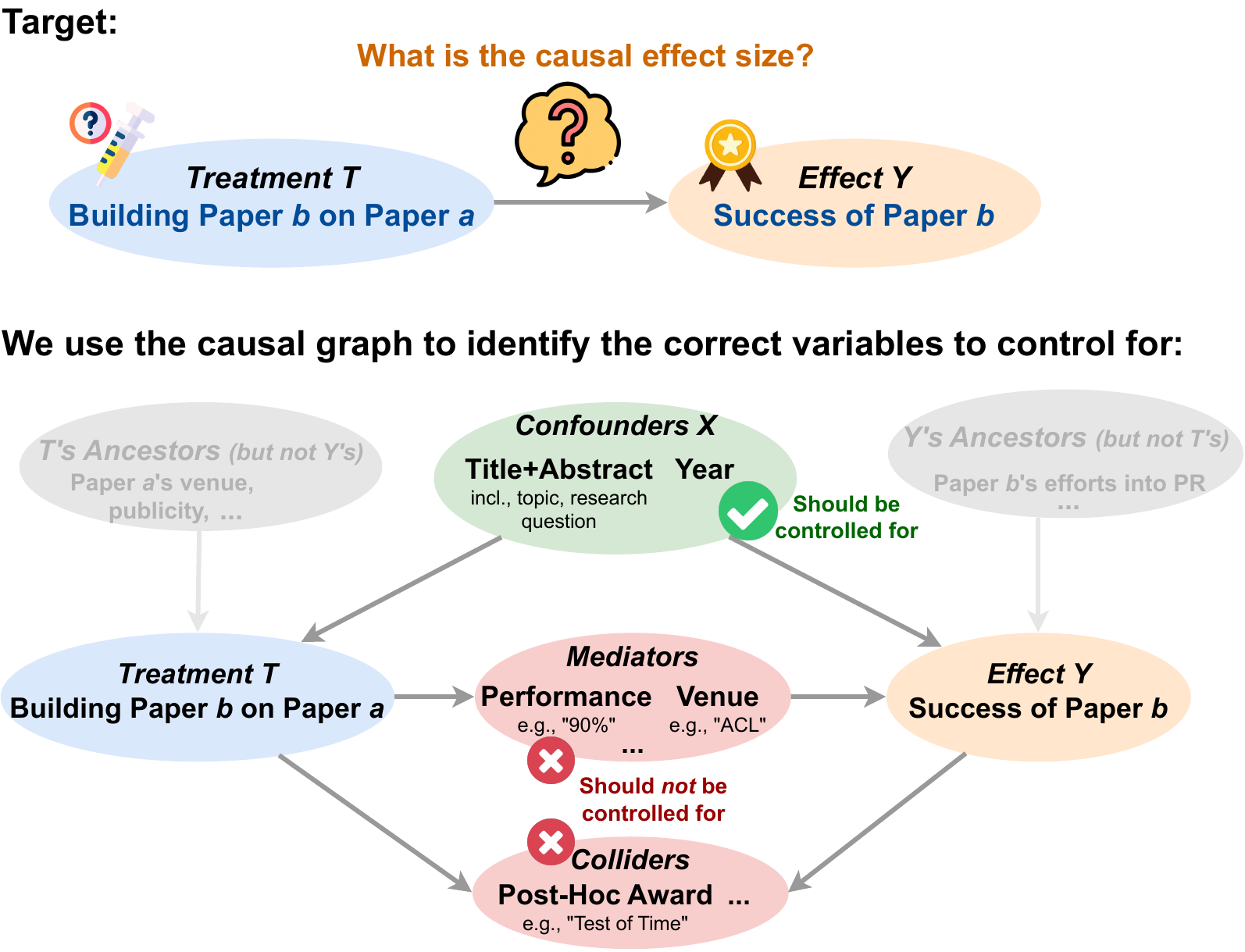}
    \caption{The causal graph of our study. 
}

    \label{fig:graph}
    \vspace{-10pt}
\end{figure}

\paragraph{Causal Graph.}
The causal graph, shown in \cref{fig:graph}, highlights the contribution of a paper $a$ to a follow-up paper $b$.
We use a binary variable $T$ to indicate if $a$ influences $b$ and an effect variable $Y$ to represent the success of $b$.
We use $\log_{10}$ of citation counts to quantify $Y$, although other transformations can also be used.
We introduce two sets of variables in this causal graph:
(i) The set of confounders, which are the common causes of $T$ and $Y$.
For instance, the research area of $b$ impacts both the likelihood of a paper citing $a$ and its own citation count.
(ii) Descendants of the treatment, comprising mediators (e.g., paper $a$ influencing the quality of paper $b$ and subsequently influencing its citations) and colliders (e.g., both the influence from $a$ and the citations of $b$ influencing later awards received by $b$).

\subsection{\ourname Indices}
In this section, we introduce various indices that measure the causal impact of a paper.

\paragraph{Two-Paper Interaction: Pairwise Causal Impact (PCI).}
To examine the causal impact of a paper $a$ on a follow-up paper $b$, we define the pairwise causal impact $\mathrm{PCI}(a, b)$ by unit-level causal effect:
\begin{align}
    \mathrm{PCI}(a, b) \coloneqq y^{t=1} - y^{t=0}
    ~,
\end{align}
where we compare the outcomes $Y$ of the paper $b$ had it been influenced by paper $a$ or not, denoted as the actual $y^{t=1}$ and the counterfactual $y^{t=0}$, respectively.
Note that the counterfactual
$y^{t=0}$ can never be observed, but only estimated by statistical methods, as we will discuss in \cref{sec:prev_method}.
\ifperfect \textcolor{gray}{
We take the commonly adopted stable unit treatment value assumption (SUTVA) \citep{rubin1980}, where there is no interaction among the units, such as peer effects for students assigned in the same class.
\fi

\paragraph{Single-Paper Quality Metrics: Total Causal Impact (TCI) and Average Causal Impact (ACI).}
Let $\bm{S}$ denote the set of all follow-up studies of paper $a$.
We define total causal impact $\mathrm{TCI}(a)$ as the sum of the pairwise causal impact index $\mathrm{PCI}(a, b)$ across all $b \in \bm{S}$.
That is,
\begin{equation}
    \mathrm{TCI}(a) \coloneqq \sum_{b \in \bm{S}} \mathrm{PCI}(a, b)~.
    \label{eq:tci}
\end{equation}
This definition provides an aggregated measure of a paper's influence across all its follow-up papers.

As the causal inference literature is usually interested in the average treatment effect, we further define the average causal impact (ACI) index as the average per paper PCI:
\begin{equation}
    \mathrm{ACI}(a) \coloneqq \frac{\mathrm{TCI}(a)}{|\bm{S}|} = \frac{1}{|\bm{S}|} \sum_{b \in \bm{S}} \left( y^{t=1} - y^{t=0}\right)
    ~.
    \label{eq:aci}
\end{equation}
\ifarxiv 
We note that $\mathrm{ACI}(a)$ is equal to the \textbf{a}verage \textbf{t}reatment effect on the \textbf{t}reated (ATT) of paper $a$ \citep{pearl2009causality}.
\fi

\section{The \ourmodel Method}\label{sec:method}

As illustrated in \cref{fig:intro}, the objective of our study is to quantify the causal effect of the treatment $T$ (i.e., whether paper $b$ is built on paper $a$) on the effect $Y$ (i.e., the outcome of paper $b$).
To approach this, we envision a counterfactual scenario: what if paper $a$ had never been published, yet certain key characteristics of paper $b$ remain unchanged?
The critical question then becomes: which key characteristics of paper $b$ should be \emph{controlled} for in this hypothetical situation?

\subsection{What Does Causal Inference Tell Us about What Variables to Control for, and What Not?}
    In causal inference, selecting the appropriate variables for control is a delicate and crucial process that affects the accuracy of the analysis.
    Pearl's seminal work on causality guides us in differentiating between various types of variables \cite{pearl2009causality}.
    
    Firstly, we must control for \emph{confounders} – variables that influence both the treatment and the outcome.
    Confounders can create spurious correlations; if not controlled, they can lead us to mistakenly attribute the effect of these external factors to the treatment itself.
    For example, in assessing the impact of one paper on another, if both papers are in a trending research area, the apparent influence might be due to the popularity of the topic rather than the papers’ content.
    
    However, not all variables warrant control.
    Mediators and colliders should be explicitly avoided in control.
    Mediators are part of the causal pathway between the treatment and outcome.
    By controlling them, we would block the very effect we are trying to measure.
    Colliders, affected by both the treatment and the outcome, can introduce bias when controlled.
    Controlling a collider can inadvertently create associations that do not naturally exist.
    In general, this also includes not controlling for the descendants of the treatment, as it could obscure the direct impact we intend to study.
    
    Lastly, variables that do not share a causal path with both the treatment and outcome, known as \emph{unshared ancestors}, are less critical in our analysis.
    They do not contribute to or confound the causal relationship we are exploring, and thus, controlling for them does not add value to our causal understanding.

\subsection{Can Existing Causal Inference Methods Handle This Control?}\label{sec:prev_method}

Several causal inference methods have been proposed to address the problem of estimating treatment effects while controlling for confounders.
Next, we will discuss the workings and limitations of three classical methods.

\paragraph{Randomized Control Trials (RCTs) Assumes Intervenability.}
The ideal way to obtain causal effects is through randomized control trials (RCTs).
For example, when testing a drug, we randomly split all patients into two groups, the control group and the treatment group, where the random splitting ensures the same distribution of the confounders across the two groups
such as gender and age.
However, RCTs are usually not easily achievable, in some cases too expensive (e.g., tracking hundreds of people's daily lives for 50 years), and in other cases unethical (e.g., forcing a random person to smoke), or infeasible (e.g., getting a time machine to change a past event in history).

For our research question on a paper's impact, utilizing RCTs is impractical as it is infeasible to randomly divide researchers into two groups, instructing one group to base their research on a specific paper $a$ while the other group does not, and then observe the citation count of their papers years later.

\paragraph{Ratio Matching Iterates over Discretized Confounder Values.}
In the absence of RCTs, matching is as an alternate method for determining causal effects from observational data.
In this case, we can let the treatment assignment happen naturally, such as taking the naturally existing set of papers and running causal inference by adjusting for the variables that block all paths.
Given a set of naturally observed papers, one of the most commonly used causal inference methods is ratio matching \citep{rosenbaum1983central}, whose basic idea is to iterate over all possible values $\bm{x}$ of the adjustment variables $\bm{X}$ and obtain the difference between the treatment group $\mathcal{T}$ and control group $\mathcal{C}$:
\begin{equation}
    \widehat{\mathrm{ACI}}(a) 
    = \sum_{\bm{x}} P(\bm{x}) \left( \frac{1}{|\mathcal{T}_{\bm{x}}|} \sum_{i \in \mathcal{T}_{\bm{x}}} y_i - \frac{1}{|\mathcal{C}_{\bm{x}}|} \sum_{j \in \mathcal{C}_{\bm{x}}} y_j
    \right)
    ~,
\end{equation}
where for each value $\bm{x}$, we extract all the units corresponding to this value in the treatment and control sets, compute the average of the effect variable $Y$ for each set, and obtain the difference.

While ratio matching is practical when there is a small set of values for the adjustment variables to sum over, its applicability dwindles with high-dimensional variables like text embeddings in our context. This scenario may generate numerous intervals to sum over, presenting numerical challenges and potential breaches of the positivity assumption.

\paragraph{One-to-One Matching Is Susceptible to Variance.}
To handle high-dimensional adjustment variables,
one possible way is to avoid pre-defining all their possible intervals, but, instead, iterating over each unit in the treatment group to match for its closest control unit \cite[e.g.,][]{mcgue2010causal,sato2022twin}.
Consider a given follow-up paper $b$, and a set of candidate control papers $\bm{C}$, where each paper $c_i$ has
a citation count $y_i$, and vector representation $\bm{t}_i$ of the confounders  (e.g., research topic).
One-to-one matching estimates PCI as
\begin{equation}
\begin{split}
\widehat{\mathrm{PCI}}(a, b) 
    &= y_b - y_{\argmax_{c_i \in \bm{C}} m_i}
     \\
    &= y_b - y_{\argmax_{c_i \in \bm{C}}\mathrm{sim}(\bm{t}_b, \bm{t}_i
)}
    ~,
\end{split}
\label{eq:match_score}
\end{equation}
where we approximate the counterfactual sample by the paper $c_i\in \bm{C}$ which is the most similar to paper $b$ by the matching score $m_i$, which is obtained by the cosine similarity $\mathrm{sim}$ of the confounder vectors.
A limitation of the
one-to-one matching method is that it
might induce large  instability in the result, as only taking one paper 
with similar contents
may have a large variance in citations when the matched paper slightly differs.

\subsection{How Do We Extending Causal Inference to Text Variables?}

\subsubsection{Theoretical Formulation of \ourmodel: Stabilizing Text Matching by Synthesis}\label{sec:method}
To fill in the aforementioned gap in the existing matching methods, we propose \ourmodel, which mitigates the instability issue of one-to-one matching by replacing it with
a convex combination of a set of matched samples to form a synthetic counterfactual sample.
Specifically, we identify a set of papers $c_i \in \bm{C}$ with high matching scores $m_i$ to the paper $b$, and synthesize the counterfactual sample by an interpolation of them:
\begin{align}
    \widehat{\mathrm{PCI}}(a,b) = y_b - \sum_{c_i \in \bm{C}} w_i y_i
     = y_b - \sum_{c_i \in \bm{C}} \frac{m_i}{\sum_{c_i \in \bm{C}} m_i} y_i
     ~,
     \label{eq:syn}
\end{align}
where the weight $w_i$ of each paper $c_i$ is proportional to the matching score $m_i$ and normalized.

The contributions of our method are as follows:
(1) we adapt the traditional matching methods from low-dimensional covariates to any high-dimensional variables such as text embeddings;
(2)
different from the ratio matching, we do not stratify the covariates, but synthesize a counterfactual sample for each observed treated units;
(3) due to this iteration over each treated unit instead of taking the population-level statistics, we closely control for exogenous  variables for the ATT estimation, which circumvents that need for 
the structural causal models;
(4) we further stabilize the estimand by a convex combination
of a set of similar papers. Note that the contribution of \cref{eq:syn} might seem to bear similarity with synthetic control \cite{abadie2003economic,abadie2010synthetic}, but they are fundamentally different, 
in that synthetic control runs on time series, and fit for the weights $w_i$ by linear regression between the time series of the treated unit and a set of time series from the control units, using each time step's values in the regression loss function.

\subsubsection{Overall Algorithm}
To operationalize our theoretical formulation above, we introduce our overall algorithm
in Algorithm~\ref{alg:getPCI}. We briefly give an overview of the the algorithm with more details to be elaborated in later sections. We use the weighted average of the matched samples following our \ourmodel method in \cref{eq:syn} through
\cref{line:eq_prep1,line:eq_prep2,line:eq_prep3,line:eq_prep4,line:eq1,line:eq2,line:eq3,line:eq4,line:eq5,line:eq_final}. In our experiments, we use the interpolation of up to top 10 matched papers. We encourage future work to explore other hyperparameter settings too. Given the PCI estimation, the main spirit of the $\textsc{GetACIandTCI}(a)$ function is to average or sum over all the follow-up studies of paper $a$, following the theoretical formulation in \cref{eq:tci,eq:aci}
and implemented in our algorithm through \cref{line:add1,line:add2,line:add3,line:add4,line:add5,line:add6}.

\begin{algorithm}[ht]
\small
    \caption{Get causal impact indices $\mathrm{ACI}$ and $\mathrm{TCI}$}
    \begin{algorithmic}[1]
        \State \textbf{Input}: Paper $a$.

        \Procedure{GetACIandTCI}{$a$}
            \State $\bm{D} \gets \mathrm{GetDesc}(a)$ \Comment{Get descendants by DFS}
            \State $\bm{B} \gets \mathrm{GetChildren}(a)$
            \State $\bm{B}' \gets \mathrm{SampleSubset}(\bm{B})$
            \Comment{See \cref{sec:sampling}}
            \State $\bm{C} \gets \mathrm{EntireSet} \backslash \{\bm{D} \cup \{a\}\}$
            \Comment{Get non-descendants}
            \State $\mathrm{ACI} \gets 0$
            \label{line:add1}
            \For {each $b_i$ in $\bm{B}'$}
            \label{line:add2}
                \State $I_i \gets \textsc{GetPCI}(a, b_i, \bm{C})$
            \label{line:add3}
                \State $\mathrm{ACI} \gets \mathrm{ACI}+ \frac{1}{|\bm{B}'|} \cdot I_i$
            \label{line:add4}
            \EndFor
            \label{line:add5}
            \State $\mathrm{TCI} \gets \mathrm{ACI} \cdot |\bm{B}|$
            \label{line:add6}
            \State \textbf{return} $\mathrm{ACI}$ and $\mathrm{TCI}$
    \EndProcedure
        \Statex
        \Procedure{GetPCI}{$a, b, \bm{C}$}
            \State $\bm{C}_{\mathrm{sameYear}} \gets \mathrm{FilterByYear}(\bm{C}, b_{\mathrm{year}})$
            \label{line:year}
            \For {each $p_i$ in $\bm{C}_{\mathrm{sameYear}} \cup \{b\}$}\label{line:remove1}
                \State $\bm{t}_i \gets \mathrm{RemoveMediator}(\mathrm{TitleAbstract}_i)$
                \label{line:remove2}
            \EndFor\label{line:remove3}
            \State $\bm{C}_{\mathrm{coarse}} \gets \mathrm{BM25}(b, \bm{C}_{\mathrm{sameYear}}, \text{topk}=100)$ \label{line:bm}
            \For {each $c_i$ in $\bm{C}_{\mathrm{coarse}}$} \label{line:sim1}
                \State $m_i \gets \mathrm{Sim}(\bm{t}_b, \bm{t}_{i})$
                \label{line:sim}
            \EndFor \label{line:sim2}
            \State $\bm{C}_{\mathrm{top10}} \gets \mathrm{argmax10}_m(\bm{C}_{\mathrm{coarse}})$ 
            \Statex
            \State $M \gets 0$ \label{line:eq_prep1}
            \For {each $c_i$ in $\bm{C}_{\mathrm{top10}}$} \Comment{For the normalization later}\label{line:eq_prep2}
                \State $M \gets M + m_i$\label{line:eq_prep3}
            \EndFor\label{line:eq_prep4}
            \State $\hat{y}^{t=0} \gets 0$ 
            \label{line:eq1}
            \For {each $c_i$ in $\bm{C}_{\mathrm{top10}}$}
            \label{line:eq2}
                \State $w_i \gets \frac{m_i}{M} $
                \label{line:eq3}
                \State $\hat{y}^{t=0} \gets \hat{y}^{t=0} + w_i \cdot y_i$
                \Comment{Apply \cref{eq:syn}}
                \label{line:eq4}
            \EndFor            \label{line:eq5}
            \State \textbf{return} $y_b - \hat{y}^{t=0}$ \label{line:eq_final}
        \EndProcedure
    \end{algorithmic}\label{alg:getPCI}
\end{algorithm}

\subsubsection{Key Challenges and Mitigation Methods}
We address several technical challenges below.

\vspace{-1em}
\subsubsubsection{Confounders of Various Types}
\vspace{-0.4em}
First, as we mentioned in the causal graph in \cref{fig:graph}, the confounder set consists of a text variable (title and abstract concatenated together) and an ordinal variable (publication year). Therefore, the similarity operation $\mathrm{Sim}$ between two papers should be customized. For our specific use case, we first filter by the publication year in \cref{line:year}, as it is not fair to compare the citations of papers published in different years. Then, we apply the cosine similarity method paper embeddings as in \cref{line:sim}. As a general solution, we recommend to separate hard logical constraints, and soft matching preferences, where the hard constraints should be imposed to filter the data first, and then all the rest of the variables can be concatenated to apply the similarity metric on.

\vspace{-1em}
\subsubsubsection{Excluding the Mediators from Confounders%
}
\label{sec:adjust}
\vspace{-0.4em}
Another key challenge to highlight is that the text variable we use for the confounder might accidentally include some mediator information. For example, the quality or performance of a paper could be expressed in the abstract, such as ``we achieved 90\% accuracy.''
Therefore, we conduct a specific preprocessing procedure before feeding the text variable to the similarity function. For the $\mathrm{RemoveMediator}$ function in \cref{line:remove2}, we exclude all numerical expressions such as percentage numbers, as well as descriptions such as ``state-of-the-art.'' For generalizability, the essence of this step is a entanglement action to separate the confounder variable (in this case, the research content) and all the descendants of the treatment variable (in this case, mentions of the performance). For more complicated cases in future work, we recommend a separate disentanglement model to be applied here.

\vspace{-1em}
\subsubsubsection{Efficient Matching-and-Reranking Method
}
\vspace{-1.5em}
Since we use one of the largest available paper databases, the Semantic Scholar dataset \citep{Kinney2023TheSS} containing 206M papers, we need to optimize our algorithm for large-scale paper matching.
For example, after we filter by the publication year, the number of candidate papers $\bm{C}_{\mathrm{sameYear}}$ could be up to 8.8M.
In order to conduct text matching across millions of papers, we use a \textit{matching-and-reranking} approach, by combining
two NLP tasks, information retrieval (IR) \citep{cosineSim} and semantic textual similarity (STS) \citep{majumder2016semantic,chandrasekaran2022evolution}. 

Specifically, we first run large-scale matching to obtain 100 candidates papers (\cref{line:bm}) using the common IR method,
BM25 \citep{bm25}.
Briefly, BM25 is a bag-of-words retrieval function that uses term frequencies and document lengths to estimate relevancy between two text documents. Deploying this method, we can find a set of candidate papers for, for example, {two} million papers, at a speed {250}x faster than the text embedding cosine similarity matching.
Then, we conduct
a fine-grained reranking using cosine similarity 
(\cref{line:sim1,line:sim,line:sim2}). In the cosine similarity matching process, 
we use the MPNet model \cite{song2020mpnet} to encode the text of each paper $c_i$ into an embedding $\bm{t}_i$, with which we get the matching score $m_i$ according to \cref{eq:match_score} in \cref{line:sim}, and the normalized weight $w_i$ by \cref{eq:syn} in \cref{line:eq3}.

\vspace{-1em}
\subsubsubsection{Numerical Estimation
}\label{sec:sampling}
Given the large number of papers, it is numerically challenging to aggregate the TCI from individual PCIs,
because the number of follow-up papers for a study can be up to tens of thousands, such as the 57,200 citations by 2023 for the ImageNet paper \citep{deng2009imagenet}.
To avoid extensively running PCI for all follow-up papers, we propose a new numerical estimation method using a carefully designed random paper subset.

A naive way to achieve this aggregation is Monte Carlo (MC) sampling. However, %
unfortunately, MC sampling requires very large sample sizes when it comes to estimating long-tailed distributions, which is the usual case of citations. Since citations are more likely to be concentrated in the head part of the distribution, we cannot afford the computational budget for huge sample sizes that cover the tails of the distribution. Instead, we propose a novel numerical estimation method for sampling the follow-up papers, inspired by importance sampling \citep{Singh2014SamplingT,Kloek1976BayesianEO}.

\ifperfect
\begin{align}
    \mathrm{TCI}(a) \coloneqq |\bm{S}| \cdot \frac{1}{N} \sum_{i=1}^{N} \mathrm{PCI}(a, b_i) 
    ~.
\end{align}
\fi

Our numerical estimation method works as follows: First, we propose the formulation that the relation between ACI and TCI is an integral over all possible paper $b$'s. Then, we formulated the above sampling problem as integral estimation or area-under-the-curve estimation. 
We draw inspiration from Simpson's method, which estimates integrals by binning the input variable into small intervals. Analogously, although we cannot run through all PCIs, we use citations as a proxy, bin the large set of follow-up papers according to their citations into $n$ equally-sized intervals, and perform random sampling over each bin, which we then sum over. In this way, we make sure that our samples come from all parts of the long-tailed distribution and are a more accurate numerical estimate for the actual TCI.

\section{Performance Evaluation}
    The contribution of a paper is inherently multi-dimensional, making it infeasible to encapsulate its richness fully through a scalar.
    Yet the demand for a single, comprehensible metric for research impact persists, fueling the continued use of traditional citations despite their known limitations.
    In this section, we show how our new metrics significantly improve upon traditional citations by providing quantitative evaluations comparing the effectiveness of citations, Semantic Scholar's highly influential (SSHI) citations \citep{semanticHICA}, and our \ourname metric.
\subsection{Experimental Setup}
\myparagraph{Dataset} We use the Semantic Scholar dataset \citep{lo-etal-2020-s2orc,Kinney2023TheSS}\footnote{\href{https://api.semanticscholar.org/api-docs/datasets}{https://api.semanticscholar.org/api-docs/datasets}} which includes a corpus of 206M scientific papers, and a citation graph of 2.4B+ citation edges. For each paper, we obtain the title and abstract for the matching process. We list some more details of the dataset in \cref{appd:data}, such as the number of papers reaching 8M per year after 2012.

\paragraph{Selecting the Text Encoder}
When projecting the text into the vector space, we need a text encoder with a strong representation power for scientific publications,
and is sensitive towards two-paper similarity comparisons regarding their abstracts containing key information such as the research topics.
For the representation power for scientific publications, instead of general-domain models such as BERT \citep{devlin-etal-2019-bert} and RoBERTa \citep{liu2019roberta}, 
we consider LLM variants\footnote{Note that we follow the standard notion by \citet{Yang2023HarnessingTP} to refer to BERT and its variants as LLMs.} pretrained on large-scale scientific text, such as SciBERT \citep{beltagy-etal-2019-scibert}, SPECTER \citep{specter2020cohan}, and MPNet \citep{song2020mpnet}.

To check the quality of two-paper similarity measures, 
we conduct a small-scale empirical study comparing human-ranked paper similarity and model-identified semantic similarity in \cref{appd:emb}, according to which MPNet outperforms the other two models.

\myparagraph{Implementation Details}\label{sec:impl}
We deploy the \textit{all-mpnet-base-v2} checkpoint of the MPNet using the \textit{transformers} Python package \citep{wolf2019transformers}, and set the batch size to be 32.
For the set of matched papers, we consider papers with cosine similarity scores higher than 0.81, which we optimize empirically on 100 random paper pairs. We the top ten most similar papers above the threshold. 
In special cases where there is no matched paper above the threshold, it means that no other paper works on the same idea as Paper $b$, and we make the counterfactual citation number to be zero, which also reflects the quality of Paper $b$ as its novelty is high.

To enable efficient operations on the large-scale citation graph, we use the Dask framework,\footnote{\href{https://dask.org/}{https://dask.org/}} which optimizes for data processing and distributed computing.
We optimize our program to take around 100GB RAM, and on average 25 minutes for each $\mathrm{PCI}(a,b)$ after matching against up to millions of candidates. More implementation details are in \cref{appd:impl}. 
For the estimation of TCI, we empirically select the sample size to be 40, which is a balance between the computational time and performance, as found in \cref{appd:sample_size}.

\subsection{Author-Identified Paper Impact}
    In this experiment, we follow the evaluation setup in \citet{semanticHICA} to use an annotated dataset \citep{zhu} comprised of 1,037 papers, annotated according to whether they serve as significant prior work for a given follow-up study.
    Although paper quality evaluation can be tricky,
    this dataset was cleverly annotated by first collecting a set of follow-up studies and letting one of the authors of each paper go through the references they cite and select the ones that significantly impact their work. In other words, for a given paper $b$, each reference $a$ is annotated as whether $a$ has significantly impacted $b$ or not.

\cref{tab:zhu} reports the accuracy of our \ourname metric, together with two existing citation metrics: citations, and SSHI citations \citep{semanticHICA}. See the detailed derivation of the accuracy scores in \cref{appd:zhu}.
From this table, we can see that our \ourname metric achieves the highest accuracy, 80.29\%, which is 5 points higher than SSHI, and 9 points higher than the traditional citations.

\subsection{Test-of-Time Paper Analysis}

\begin{figure*}[th!]
  \begin{minipage}{0.3\textwidth}
    \centering
    \small 
    \centering
\begin{tabular}{lcl}
            \toprule
            \textbf{Metric}  & \textbf{Accuracy} \\
            \midrule
            Citations & 71.33  \\ 
            SSHI Citations & 75.25\\
            \ourname  & \textbf{80.29} \\ 
            \bottomrule
        \end{tabular}
        \captionof{table}{Accuracy of all three citation metrics.
        }
        \label{tab:zhu}
        
    \begin{tabular}{lccccll}
        \toprule
        \textbf{Metric}  & \textbf{Corr. Coef.} \\
        \midrule 
        Citations & 0.491  %
        \\ 
        SSHI Citations & 0.317\\
        TCI       & \textbf{0.640} %
        \\
        \bottomrule
    \end{tabular}
    \label{tab:fullComparison}
    \captionof{table}{Correlation coefficients of each metric and ToT paper award by Point Biserial Correlation \citep{PointBiserial}.
    }\label{tab:tot}
  \end{minipage}\hfill
  \begin{minipage}{0.24\textwidth}
    \centering
    
        \includegraphics[width=\linewidth]{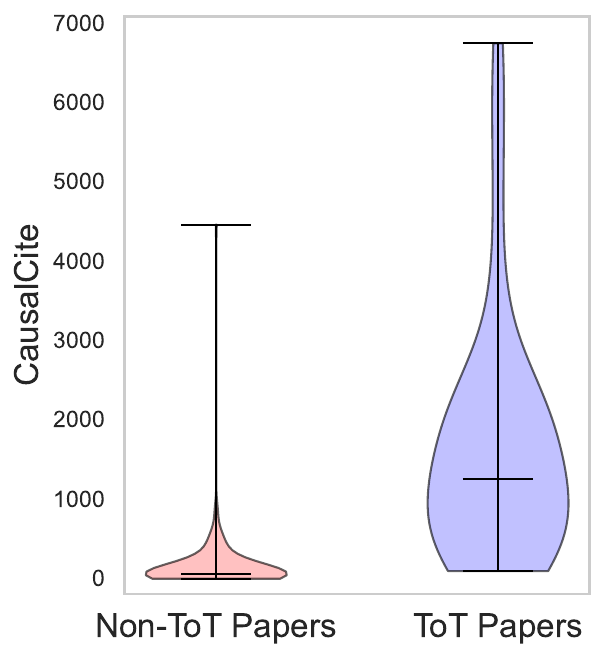}
        \caption{Distributions of ToT (mean: 142) and non-ToT papers (mean: 1,623). 
        }
        \label{fig:ToTViolin} \label{fig:tot}

  \end{minipage}\hfill
  \begin{minipage}{0.4\textwidth}
    \centering
    \includegraphics[width=\linewidth]{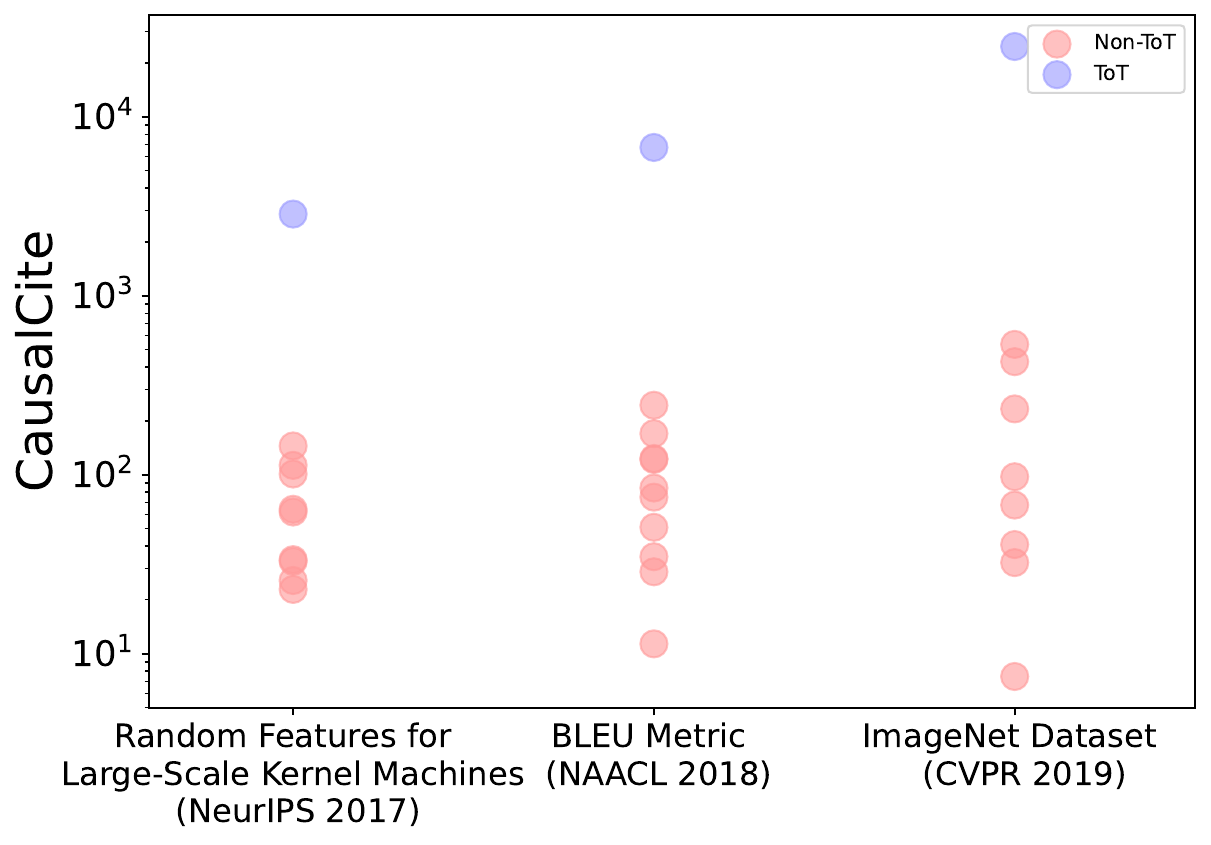}
        \caption{The \ourname values of three example ToT papers from general AI, NLP, and CV. 
        }
        \label{fig:ToTMain}\label{fig:totexample}
  \end{minipage}\hfill
\end{figure*}

    The test-of-time (ToT) paper award is a prestigious honor bestowed upon papers that have made substantial and enduring impacts in their field.
    In this section, we collect a dataset of {$792$} papers, including $72$ ToT papers, and a control group of $10$ randomly selected non-ToT papers from the same conference and year as each ToT paper.
    To collect this ToT paper dataset, we look into ten leading AI conferences spanning general AI (NeurIPS, ICLR, ICML, and AAAI), NLP (ACL, EMNLP, and NAACL), and CV (CVPR, ECCV, and ICCV), for which we go through each of their websites to identify all available ToT papers.\footnote{We get this list by selecting the top conferences on Google Scholar using the h5-Index ranking in each of the above domains: general AI (\href{https://scholar.google.com/citations?view_op=top_venues&vq=eng_artificialintelligence}{link}), CV (\href{https://scholar.google.com/citations?view_op=top_venues&vq=eng_computervisionpatternrecognition}{link}), and NLP (\href{https://scholar.google.com/citations?view_op=top_venues&vq=eng_computationallinguistics}{link}).}

In \cref{tab:tot}, we show the correlations of various metrics with the ToT awards.
    In this table, \ourname achieves the highest correlation of 0.639, which is +30.14\% better than that of citations.
    Furthermore, we visualize the correspondence of our metric and ToT, and observe a substantial difference between
    the
    \ourname distributions of ToT vs. non-ToT papers
    in \cref{fig:ToTViolin}. We also show three examples of ToT papers in \cref{fig:totexample}, where the ToT papers differ from the non-ToT papers by one or two orders of magnitude.

\subsection{Topic Invariance of \ourname}
    \begin{table}[ht]
        \small 
        \centering
        \begin{tabular}{lcccccc}
            \toprule
            \textbf{Research Area}
            & \textbf{\pciavg} & \textbf{Citations} & \textbf{SSHI} \\
            \midrule
            General AI (n=16) & 0.748
            & 2,024  & 267\\
            CV (n=36) & 0.734
            & 7,238  & 1,088  \\
            NLP (n=20)    & 0.763 
            & 1,785 & 461    \\ 
            \bottomrule
        \end{tabular}
        \caption{The average of each metric by research area on our collected set of 72 ToT papers.}
\label{tab:topicAgnostic}
    \end{table}

A well-known issue with citations is their inconsistency across different fields.
What might be considered a large number of citations in one field might be seen as average in another.
In contrast, we show that our ACI index does not suffer from this issue.
We show this using our ToT dataset, where we control for the quality of the papers to be ToT but vary the domain by the three fields: general AI, CV, and NLP.
We observe in \cref{tab:topicAgnostic} that even though some domains have significantly more citations (for instance, CV ToT papers have, on average, $4.05$ times more citations than NLP), the ACI remains consistent across various fields.

\section{Findings}
Having demonstrated the effectiveness of our metrics, we now explore some open-ended questions:
(1) Do best papers have high causal impact? (\cref{sec:best})
(2) How does the \ourname value distribute across papers?
(\cref{sec:curve}) 
(3) What is the impact of some famous papers evaluated by \ourname? (\cref{sec:famous_papers})
(4) Can we use this metric to correct for citations? (\cref{sec:outlier}).

\ifperfect \zhijing{Reviewer 2 of ICLR suggests that we have a stronger head-to-head comparison with citations, highlighting (1) criticisms of the existing citation-count system, (2) How does CAUSALCITE solve the claimed limitations.}
\fi

\subsection{Do Best Papers Have High Causal Impact?} \label{sec:best}
Selecting best paper awards is an arguably much harder task than ToT papers, as it is difficult to predict of the impact of a paper when it is just newly published.
Therefore, we are interested in the actual causal impact of best papers. Similar to our study on ToT papers, we collect a dataset of {$444$} papers including $74$ best papers and a control set of random $5$ non-best papers from the same conference in the same year, using the same set of the top ten leading AI conferences.
We find that the correlation of the \ourname metric with best papers is $0.348$, which is very low compared to the $0.639$ correlation with the ToT papers.
This shows that the best papers do not necessarily have a high causal impact.
One interpretation can be that the best paper evaluation is a forecasting task, which is much more challenging than the retrospective task of ToT paper selection.

\subsection{What Is the Nature of the \ourname Distribution?}\label{sec:curve}
\begin{figure}[ht]
    \centering
    \includegraphics[width=.9\linewidth]{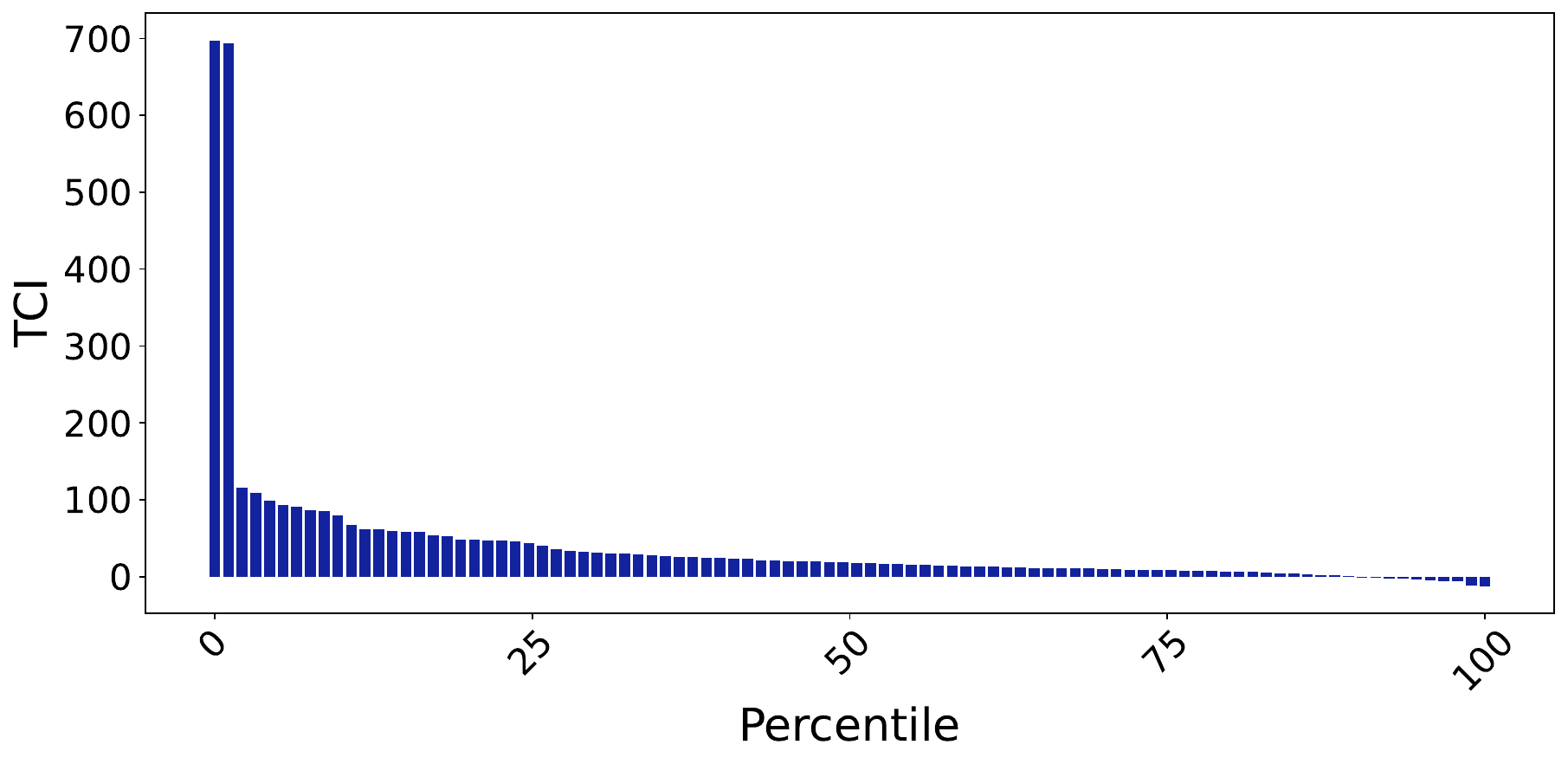}
        \caption{The distribution of TCI values by percentile of 100 random papers, which shows a long tail indicating that high impact is concentrated in a relatively small portion of papers.
        }
    \label{fig:curve}
\end{figure}

We explore how the \ourname scores are distributed across papers in general. We plot \cref{fig:curve} using a random set of 100 papers from the Semantic Scholar dataset, which is a reasonably large size given the computation budget mentioned in \cref{sec:impl}.
From this plot, we can see 
a power law distribution with a long tail, echoing with the common belief that the paper impact follows the power law, with high impact concentrated in a relatively small portion of papers.

\subsection{Selected Paper Case Study} \label{sec:famous_papers}
\begin{table}[ht]
    \small 
    \centering
        \begin{tabular}{lccc}
            \toprule
            Paper Name & TCI    & Citations & \pciavg \\ \midrule
            Transformers  & 52,507 & 68,064    & 0.771                         \\
            BERT       & 40,675 & 59,486    & 0.683                         \\
            RoBERTa    & 6,932  & 14,434    & 0.480                         \\
            
            \bottomrule
        \end{tabular}
        \caption{Case study of some selected NLP
        papers.}
        \label{tab:empirical}

\end{table}    
In addition to the shape of the overall distribution, we also look at our metric's correspondence to some selected papers shown in
\cref{tab:empirical}. For example, we know that the Transformer paper \citep{vaswani2017attention} is a more foundational work than
its follow-up work BERT \citep{devlin-etal-2019-bert}, and BERT is more foundational than its later variant, RoBERTa \citep{liu2019roberta}. This monotonic trend is confirmed in their TCI and ACI values too. Again, this is a preliminary case study, and we welcome future work to cover more papers.

\subsection{Discovering Quality Papers beyond
Citations}
\label{sec:outlier}

Another important contribution of our metric is that it can help discover papers that are traditionally overlooked by citations.
To achieve the discovery, we formulate the problem as outlier detection, where we first use a linear projection to handle the trivial alignment of citations and \ourname,
and then analyze the outliers using the interquartile range (IQR) method \citep{outliers}. See the exact calculation in \cref{appd:outlier}.
We show the three subsets of papers in \cref{tab:outlier}, where the two outlier categories, the overcited and undercited papers, correspond to the false positive and false negative oversight by citations, respectively. 
An additional note is that, when we look into some characteristics of the three categories, we find that the citation frequency in result section, i.e., the percentage of times they are cited in results section compared to all the citations, correlates with these categories.
Specifically, we find that the undercited papers tend to have more of their citations concentrated in the results section, which usually indicates that this paper constitutes an important baseline for a follow-up study, while the overcited papers tend to be cited out of the results section, which tends to imply a less significant citation.

\begin{table}[ht]
\small 
\centering
\begin{tabular}{lcccccc}
\toprule
Paper Category & Result Citations & Residual %
\\ \midrule

Overcited Papers ({7.04}\%) & 1.26 & -1.792 \\
Aligned Papers ({91.20}\%) & 1.51 & 0.118 \\
Undercited Papers ({1.76}\%) & 1.90  & 1.047 \\
\bottomrule
\end{tabular}
\caption{We use our \ourname metric to discover outlier papers that are overlooked by citations. 
For each paper category, we include their portion relative to the entire population, the percentage of citations occurred in the result section (Result Citations), and average residual value by linear regression.
}\label{tab:outlier}
\end{table}

\section{Related Work}
    The quantification of scientific impact has a rich history and continuously evolves with technology.
    Bibliometric analysis has been largely influenced by early methods that relied on citation counts \citep{garfield1964use, garfield1972citation, garfield1964science}.
    \citet{hou2017exploration} investigate the evolution of citation analysis, employing reference publication year spectroscopy (RPYS) to trace its historical development in scientometrics.
    \citet{donthu2021conduct} provide practical guidelines for conducting bibliometric analysis, focusing on robust methodologies to analyze scientific data and identify emerging research trends.

    Indices such as the h-index, introduced by \citet{hirsch2005index}, are established tools for measuring research impact.
    The more recent Relative Citation Ratio (RCR), developed by \citet{hutchins2016relative}, provides a field-normalized alternative to traditional metrics.
    \citet{semanticHICA} introduced SSHI, an approach to identify meaningful citations in scholarly literature.
    However, these metrics are not without limitations.
    As \citet{wroblewska2021research} discussed, conventional citation-based metrics often fail to capture the multidimensional nature of research impact.
    In this context, \citet{elmore2018altmetric} discussed the Altmetric Attention Score, which evaluates the broader societal and online impact of research.
    
    With the increasing availability of large datasets and the advent of digital technologies, new opportunities for bibliometric analysis have emerged.
    \citet{iqbal2021decade} highlighted the role of NLP and machine learning in enhancing in-text citation analysis.
    Similarly, \citet{umer2021scientific} explored the use of textual features and SMOTE resampling techniques in scientific paper citation analysis.
    \citet{jebari2021use} analyzed citation context to detect research topic evolution, showcasing data analysis for scientific discourse.
    \citet{chang2023citesee} explored augmenting citations in scientific papers with historical context, offering a novel perspective on citation analysis.
    \citet{manghi2021new} introduced scientific knowledge graphs, an innovative method for evaluating research impact.
    \citet{bittmann2021applied} explored statistical matching in bibliometrics, discussing its utility and challenges in post-matching analysis.
    The use of AI in bibliometric analysis is highlighted in research by \citet{chubb2022speeding} and the systematic review of AI in information systems by \citet{collins2021artificial}.
    Network analysis approaches, as discussed by \citet{chakraborty2020patent} in the context of patent citations and by \citet{dawson2014current} in learning analytics, further illustrate the diverse applications of advanced methodologies in understanding citation patterns.

\ifperfect

Citet papers:

\textbf{Citation Analysis Techniques and Metrics}
A decade of in-text citation analysis based on natural language processing and machine learning techniques: an overview of empirical studies \citep{iqbal2021decade} \\
How to conduct a bibliometric analysis: An overview and guidelines \citep{donthu2021conduct}\\
Scientific papers citation analysis using textual features and SMOTE resampling techniques \citep{umer2021scientific} \\
Research impact evaluation and academic discourse \citep{wroblewska2021research}\\
New trends in scientific knowledge graphs and research impact assessment \citep{manghi2021new} \\
Exploration into the evolution and historical roots of citation analysis by referenced publication year spectroscopy \citep{hou2017exploration}\\
The use of citation context to detect the evolution of research topics: a large-scale analysis \citep{jebari2021use}\\
Relative citation ratio (RCR): a new metric that uses citation rates to measure influence at the article level \citep{hutchins2016relative}\\
The Altmetric attention score: what does it mean and why should I care? \citep{elmore2018altmetric}\\
Applied usage and performance of statistical matching in bibliometrics: The comparison of milestone and regular papers with multiple measurements of disruptiveness as an empirical example \citep{bittmann2021applied}\\ 
Speeding up to keep up: exploring the use of AI in the research process \cite{chubb2022speeding} \\
Artificial intelligence in information systems research: A systematic literature review and research agenda \citep{collins2021artificial} \\ 
Patent citation network analysis: A perspective from descriptive statistics and ERGMs \citep{chakraborty2020patent}\\ 
Current state and future trends: A citation network analysis of the learning analytics field \citep{dawson2014current} \\
CiteSee: Augmenting Citations in Scientific Papers with Persistent and Personalized Historical Context \citep{chang2023citesee} \\

\subsection{citation analysis}

Citations are important for direct research impact and quality indicator \citep{Aksnes},
And as performance indicators for decisions such as
tenure evaluation, grant assessment, evaluate to the impact of grants.
use of citation indicators in evaluation of the scientific performance of research groups, departments, and institutions (Moed, 2005); evaluation of research proposals (Cabezas-Clavijo, Robinson-Garcia, Escabias, Jimenez-Contreras, 2013); allocation of research funding (Carlsson, 2009); and hiring of academic personnel (Holden, Rosenberg, Barker, 2005). 

Pros and cons:
Arguably, citations might reflect scientific impact and relevance -Aksnes

acknowledged flaws, solidity/plausibility, originality, and societal value

An important study might be underestimated because it was
not published.
Simply counting citations, meanwhile, fails to capture the idea that articles should be judged relative to similar papers: an algebra paper with a few dozen citations, for example, may have a greater impact in mathematics than a widely cited cancer study would have in oncology

https://arxiv.org/abs/2305.12920

\paragraph{Matching, Synthetic Control, ...}

\paragraph{Network Annotation}
network interference (JHU)

\subsection{Different Measures}

Scholar-centric indices:

- Total citations

- h-index

Paper-centric indices:

- Citations

- Highly influential citations

Paper-to-paper indices:

- Whether the paper-to-paper citation is a highly influential citation (binary)

\subsection{Direct Citations}
\subsection{Highly Influential Citations}
\subsection{Shortcomings of Existing Metrics}

\fi
\section{Conclusion}
In this study, we propose \ourname, a novel causal formulation for paper citations. Our method combines traditional causal inference methods with the recent advancement of NLP in LLMs to provide a new causal outlook on paper impact by answering the causal question: ''Had this paper never been published, what would be the impact on this paper’s current follow-up studies?''.
With extensive experiments and analyses using expert ratings and test-of-time papers as criteria for impact, our new \ourname metric demonstrates clear improvements over the traditional citation metrics. Finally, we use this metric to investigate several open-ended questions like ``Do best papers have high causal impact?'', conduct a case study of famous papers, and suggest future usage of our metric for discovering good papers less recognized by citations for the scientific community.

\section*{Limitations and Future Work}
There are several limitations for our work. For example, as mentioned previously, our metric has a high computational budget. %
Future work can explore more efficient optimization methods. Also, we model the content of the paper by its title and abstract, it could also be possible for future work to benefit from modeling the full text, given appropriate license permissions.

As for another limitation, our study is based on data provided by the Semantic Scholar corpus. This corpora has certain properties such as being more comprehensive with computer science papers, but less so in other disciplines. Its citation data also has a delay compared to Google Scholar, so for the newest papers, the citation score may not be accurate, making it more difficult to calculate our metric.

Additionally, our study provides a general framework for causal inference given a causal graph that involves text. It is totally possible that for a more fine-grained problem, the causal graph will change, in which case, we undersuggest future researchers to derive the new backdoor adjustment set, and then adjust the algorithm accordingly. An example of such a variable could be the author information, which might also be a confounder. 

Finally, since quality evaluation of a paper is a multi-faceted task, theoretically, a single number can never give more than a rough approximation, because it collapses multiple dimensions into one and loses information. Our argument in this paper is just to show that our formulation is theoretically more accurate than the citation formulation. We take one step further, instead of solving the quality evaluation problem which is much more nuanced. Some intrinsic problems in citations that we can also not solve (because our metrics still rely on using citations, just contrasting them in the right away) include (1) if a paper is newly published, with zero citations, there is no way to obtain a positive causal index, and (2) we do not solve the fair attribution problem when multiple authors share credit of a paper, as our metric is not sensitive towards authors.

\section*{Ethical Considerations}

\textbf{Data Collection and Privacy}
The data used in this work are all from Open Source Semantic Scholar data, with no user privacy concerns. The potential use of this work is for finding papers that are unique and innovative but do not get enough citations due to loack of popularity or awareness of the field. This metric can act as an aid when deciding impact of papers, but we do not suggest its usage without expert involvement. Through this work, we are not trying to demean or criticize anyone's work we only intend to find more papers that have made a valuable contribution to the field.

\textbf{CS-Centric Perspective}
The authors of this paper work in Computer Science (mostly Machine Learning) hence a lot of analysis done on the quality of papers that required sanity checks are done on ML papers. The conferences selected for doing the ToT evaluation were also CS Top conferences, hence they might have induced some biases. The metric in general has been created generically and should be applicable to other domains as well, the Author Identified Most Influential Papers study is also done on a generalized dataset, but we encourage readers in other disciplines to try out the metric on papers from their field.

\ifarxiv
\section*{Author Contributions}\label{sec:contributions}

This project originates as part of the \textit{AI Scholar} series of projects that \textbf{Zhijing Jin} started since 2021, as she identified that causal inference over papers is a valuable research setting with sufficient data and rich causal phenomena. \textbf{Bernhard Schölkopf} came up with the formulation that the action of citation itself has a causal nature, and can thus be formulated as a causal inference question. Zhijing, Bernhard, and \textbf{Siyuan Guo} settled down the overall project design.

After the initial idea formulation, \textbf{Ishan Kumar} and {Zhijing Jin} operationalized the entire project, with vast efforts in identifying the data source; improving the theoretical formulation (together with \textbf{Ehsan Mokhtarian}, and Bernhard%
); speeding up the code efficiency; designing the evaluation and analysis protocols (with the insightful supervision from \textbf{Mrinmaya Sachan} and Bernhard, and suggestions from Siyuan); and implementing all the evaluations (with the help of \textbf{Yuen Chen}).
In the writing stage, Mrinmaya gave substantial guidance to structure the storyline of the paper, 
and Zhijing, Ehsan, Ishan, and Mrinmaya contributed significantly to the writing, with various help and suggestions from all the other authors.

\section*{Acknowledgment}
During the idea formulation stage, we are grateful for the research discussions with Kun Zhang on the vision of the AI Scholar series of projects.
During the implementation of our paper, 
we thank Zhiheng Lyu for his suggestions on efficient computer algorithm over massive graphs and large data. We also thank labmates from Max Planck Institute for constructive feedback and help on data annotation.
We thank Vincent Berenz, Felix Leeb, and Luigi Gresele for their generous support with computation resources.

This material is based in part upon works supported by the German Federal Ministry of Education and Research (BMBF): Tübingen AI Center, FKZ: 01IS18039B; by the Machine Learning Cluster of Excellence, EXC number 2064/1 – Project number 390727645; 
by the John Templeton Foundation (grant \#61156); by a Responsible AI grant by the Haslerstiftung; and an ETH Grant (ETH-19 21-1).
Zhijing Jin is supported by PhD fellowships from the Future of Life Institute and Open Philanthropy, as well as the travel support from ELISE (GA no 951847) for the ELLIS program. 
\fi

\bibliographystyle{acl_natbib}
\bibliography{sec/refs_acl,sec/refs_causality,sec/refs_zhijing,sec/refs_ai_safety,sec/refs_cogsci,sec/refs_semantic_scholar,sec/refs_nlp4sg,refs_causalcite}

\clearpage
\begin{center}
    {\Large \textbf{Appendix}}
\end{center}
\appendix

\ifperfect

\section{More Method}
\subsection{assumptions}
Below are some assumptions  \citep{rubin1980, roberts2020adjusting}.

\begin{assumption}
[Stable Unit Treatment Value Assumption (SUTVA) \citep{rubin1980}]\
For all individuals $i$, $S_i(T) = S_i(T_i)$
\end{assumption}

\begin{assumption}[Conditional Ignorability]
For all individuals $i$,
$T_i \independent S_{i}(T=1), S_{i}(T=0) \vert \bf{Z}$, where $\bf{Z}$ is the set of confounders.
In the language of \it{SCM}, this assumption implies no unobserved confounders that affect both the treatment assignment and the outcome.
\end{assumption}

\begin{assumption}[Positivity]
For all individuals i, $P(T_i = t) > 0$ for all treatments $t$. 
\end{assumption}

\subsection{More Algorithm}\label{sec:alg}

\begin{algorithm}[ht]
    \caption{Algorithms for RSCI and ITCI}
    \label{alg:overview}
    \begin{algorithmic}[1]
        \State \textbf{Input}: Paper $X$
        \State $P \gets $ All its children, namely the receiver paper set
        \State $Q \gets$ Get all its non-descent paper set Q
        \Statex
        \hrulefill
        \Procedure{get\_RSCI}{$X, Y$}
            \State $year_Y \gets$ Publication year of paper $Y$
            \State $s_Y \gets$ Citations of $Y$
            \State $t_Y \gets$ Text embedding of $Y$'s title and abstract
            \State $Q' \gets$ All the papers in $Q$ which share the same $year_Y$ \label{alg:year_filter}
            \State $total\_sim \gets 0$
            \For {each $q_i$ in $Q'$}
                \State $s_i \gets$ Citations of $q_i$
                \State $t_i \gets$ Embedding of $q_i$'s title and abstract
                \State $sim_i \gets$ Cosine similarity of $t_i$ and $t_Y$
                \State $total\_sim \gets total\_sim + sim_i$
            \EndFor
            \State $control \gets 0$
            \For{$q_i$ in $Q'$}
                \State $w_i \gets \frac{sim_i}{total\_sim}$
                \State $control \gets control + w_i \cdot s_i$
            \EndFor
            \State \textbf{return} $s_Y - control$
        \EndProcedure
        \Statex    
        \hrulefill
        \Procedure{get\_ACE}{$X$}
            \State $ACE \gets 0$
            \For {each $p_i$ in $P$}
                \State $I_i \gets \text{get\_RSCI}(X, p_i)$
                \State $ACE \gets ACE + \frac{1}{|P|} \cdot I_i$
            \EndFor
            \State \textbf{return} $ACE$
        \EndProcedure
    \end{algorithmic}
\end{algorithm}

\begin{algorithm}[ht]
    \caption{Algorithm for \ourname}\label{alg:two}
    \begin{algorithmic}[1]
        \For{$paperA$ in allPapers}
            \If{numChildrenOfPaperA != 0}
                \State paperB = getRandomChild (children\_of\_paperA)
                \State candidatePool = df [ df [ "year" ] == paperB\_year]
                \State get\_closest\_papers = BM25( candidatePool, paperB)
                \State encoded\_paperB = model (paperB\_abstract)
                \For{\_candidate\_paper in get\_closest\_papers}
                    \State descendants\_of\_paperA = BFS(CitationGraph, paperA)
                    \If{\_candidate\_paper not in descendants\_of\_paperA}
                        \State encoded\_candidate\_paper = model(\_candidate\_paper)
                        \State similarity\_score = cosineSimilarity(encoded\_paperB, encoded\_candidate\_paper)
                        \If{similarity\_score > threshold}
                            \State finalSimilarPaperPool.append([similarity\_score, citation\_count]) 
                        \EndIf   
                        \State save that
                    \EndIf
                \EndFor
            \EndIf
        \EndFor
    \end{algorithmic}
\end{algorithm}

\subsection{Simpsons curve}
\fi
\section{Additional Implementation Details}
\subsection{Time and Space Complexity Details}\label{appd:impl}

For the time cost of running the causal impact indices, each $\mathrm{PCI}(a,b)$ takes around 1,500 seconds, or 25 minutes. Multiplying this by 40 samples per paper $a$, we spend 16.67 hours to calculate each ACI or TCI for the paper's overall impact.
For a fine-grained division into the time cost, the majority of the time is spend on the BM25 indexing (800s) and the sentence embedding cosine similarities calculation (400s). The rest of the time-consuming steps are the BFS search (150-200s every time) to identify descendants and non-descendants of a paper.

For the space complexity, we loaded the 2.4B edges of the citation graph into a  parquet gzip format for faster loading, and use Dask's lazy load operation to load it part by part to RAM for better parallelization. 
The program can fit into different sizes of RAMs by modifying the number of partitions and reducing the number of workers in Dask, at the cost of an increased computation time. 
On the hard disk, citation graph takes up 19G space, and paper data takes 11G.

\subsection{Numerical Estimation Method: Finding the Sample Size}\label{appd:sample_size}

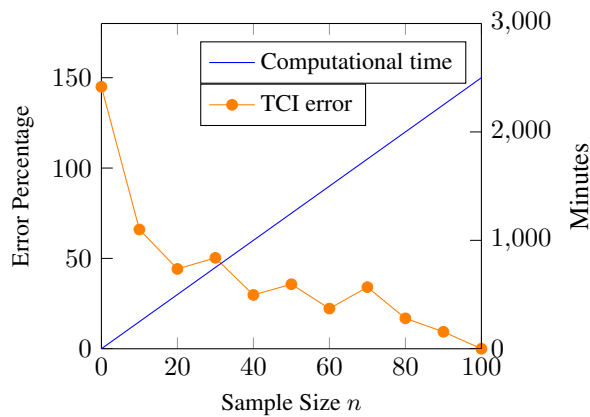
\begin{figure}[ht]
    \centering
    \begin{tikzpicture}
    \pgfplotsset{
        scale only axis,
        xmin=0, xmax=100,
        y axis style/.style={
            yticklabel style=#1,
            ylabel style=#1,
            y axis line style=#1,
            ytick style=#1
       }
    }
    
    \begin{axis}[
      axis y line*=left,
      y axis style=black!75!black,
      ymin=0, ymax=180,
      xlabel={Sample Size $n$
      },
      ylabel={Error Percentage},
    legend style={at={(0.26,0.75)}, anchor=west, font=\footnotesize},
    label style={font=\footnotesize}
    ]
    \addplot[mark=*,orange] 
   coordinates {
    (0, 144.99014469563542)
 (10, 66.00385767461768)
 (20, 44.17677805641458)
 (30, 50.314017672941745)
 (40, 29.788802957937282)
 (50, 35.71244075970475)
 (60, 22.235910683478743)
 (70, 34.097782499124584)
 (80, 16.845642879961183)
 (90, 9.392781270621153)
 (100, 0.0)
};
\addlegendentry{TCI error}
    \end{axis}
    
    \begin{axis}[
      axis y line*=right,
      axis x line=none,
      ymin=0, ymax=3000,
      ylabel={Minutes},
      y axis style=black!75!black,
      legend style={at={(0.26,0.88)}, anchor=west, font=\footnotesize}
    ]
    \addplot[color=blue, domain=0:120]{25*x};
    \addlegendentry{Computational time}
    \end{axis}
    
    \end{tikzpicture}
    \caption{
    We show the trade-off of two curves: the error curve (orange), and the time cost curve (blue). For the error curve, we see an elbow point at around $n=40$, when the error starts to be small. The curve for the computational time is linear, taking 25 minutes for each paper. Balancing the trade-offs, we decided to choose the sample size $n=40$.
        }
    \label{fig:simpson}
\end{figure}

For our numerical estimation method, we first calculate the ACI on a subset of carefully sampled papers and then aggregate it to TCI. One design choice question is how to
decide the size of this random subset. In our case, we need to balance both the computation time (25 minutes per pairwise paper impact) and the estimation accuracy.
To identify the best sample size,
we conduct a small-scale study, first obtaining the TCI using our upper-bound budget of $n=100$ samples and then gradually decreasing the number of samples to see if there is a stable point in the middle which also leads to a result close to that obtained with 100 samples.
In \cref{fig:simpson}, we show the trade-off of the two curves, the error curve and time cost,
where we can see $n=40$ seems to be a good point balancing the two. It is at the elbow of the arrow curve, making it relatively close to the estimation result of $n=100$, and also in the meantime vastly saving our computational budget, enabling us to run efficient experiments for more analyses.

\subsection{Experiment to Select the Best Embedding Method}\label{appd:emb}
When selecting the text encoder for our \ourmodel method,
we compare among the three LLMs pre-trained on scientific papers, SciBERT, MPNet, and SPECTER. Specifically, we conduct a small-scale experiment to see how much the similarities scores based on the embedding of each model align with human annotations.
As for the annotation process, we first collect a set of random papers, and for each such paper (which we call a pivot paper), we identify ten papers, from the most similar to the least, with monotonically decreasing similarity. We collect a total of {100} papers consisting of {ten} such collections, for which we show an example in \cref{tab:emb}.
Then we see how the resulting similarity scores conform to this order by deducting the percentage of papers that are out of place in the ranking. 

We find that MPNet correlates the best with human judgments, achieving an accuracy of 82\%, which is 10 points better the second best one, SPECTER, which gets 72\%, and 18 points better than SciBERT with a score of 64\%. 
It also gives more distinct scores to papers with different levels of similarity.
This capability advantage
may be attributed to its Siamese network objectives in the training process \citep{song2020mpnet}.
We open-sourced our annotated data in the codebase.

\ifperfect
In order to compare embeddings for our task we take a pivot paper and a sample of papers with decreasing levels of similarity between them. One example of such a comparison table is given in \cref{tab:emb}. We find that SciBERT doesn’t perform very well for this sentence-level embedding task as we see all the similarity scores lie within a range and we don’t see a decreasing trend in the similarity scores. SPECTER and MPNet were very close in performance, but we found that the similarity score between the pivot paper and a completely different paper (Paper 10) was a magnitude lower in the case of MPNet whereas it was only 0.1 lower in SPECTER. Due to this reason, we decided to go with MPNet as our embedding model since having a greater distance between unrelated papers is critical to our use case.
\fi
\begin{table*}[t]
\small \centering
\begin{tabular}{llccccc}
\toprule
Paper Index & Title                       & SciBERT & SPECTER & MPNet \\  \midrule
\multicolumn{2}{c}{\textit{Pivot Paper: GPT-3 \citep{Brown2020LanguageMA}}} \\
1 (Most similar)                         & PaLM \citep{Chowdhery2022PaLMSL}                        & 0.9787      & 0.8689  & 0.7679  \\
2                          & GPT-2 \citep{Radford2019LanguageMA}                        & 0.9346      & 0.9064  & 0.8196  \\
3                          & GPT \citep{Radford2018ImprovingLU}                        & 0.9488       & 0.8778  & 0.7790  \\
4                          & BERT   \citep{devlin-etal-2019-bert}                     & 0.9430      & 0.8321   & 0.6784  \\
5                          & Transformers  \citep{vaswani2017attention}              & 0.9202      & 0.8644  & 0.6385  \\
6                          & SciBERT  \citep{beltagy-etal-2019-scibert}                   & 0.8396      & 0.8112  & 0.5667  \\
7                          & Latent Diffusion Models \citep{Rombach2021HighResolutionIS}    & 0.9586      & 0.7755  & 0.4567  \\
8                          & Sentiment Analysis Using DL \citep{Fang2015SentimentAU}    & 0.7775      & 0.7298  & 0.2911  \\
9                          & Sentiment Analysis Using ML \citep{Zainuddin2014SentimentAU} & 0.6462      & 0.6403  & 0.2563  \\
10 (Least similar)                        & New High Energy Accelerator \citep{Courant1952THESS} & 0.8033     & 0.5617  & 0.0359  \\
\bottomrule

\end{tabular}
\caption{
An example collection of papers with monotonically decreasing similarity to the pivot paper. As can be seen from the similarities scores produced by the three text embedding methods, MPNet corresponds to the ground truth the most, and also shows clear score distinctions between less similar and more similar papers.
}
\label{tab:emb}
\end{table*}

\section{Dataset Overview}\label{appd:data}

    \begin{figure}[ht]
        \centering
        \includegraphics[width=\linewidth]{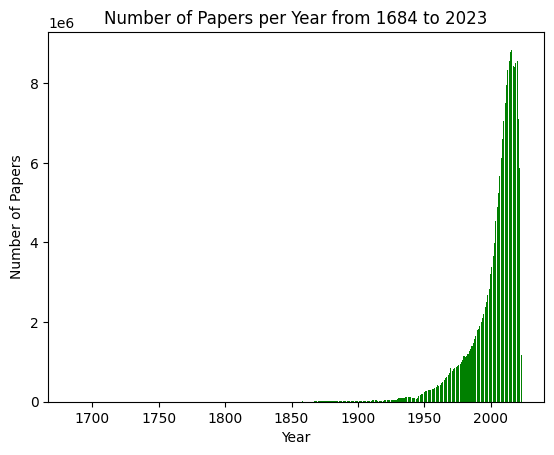}
        \caption{The 
        number of papers published per year
        from 1684 to 2023. We can see that in recent years since 2010, there are more than 7 million papers each year.}
        \label{fig:year_distr}
    \end{figure}
    \begin{figure}[ht]
        \centering
        \includegraphics[width=\linewidth]{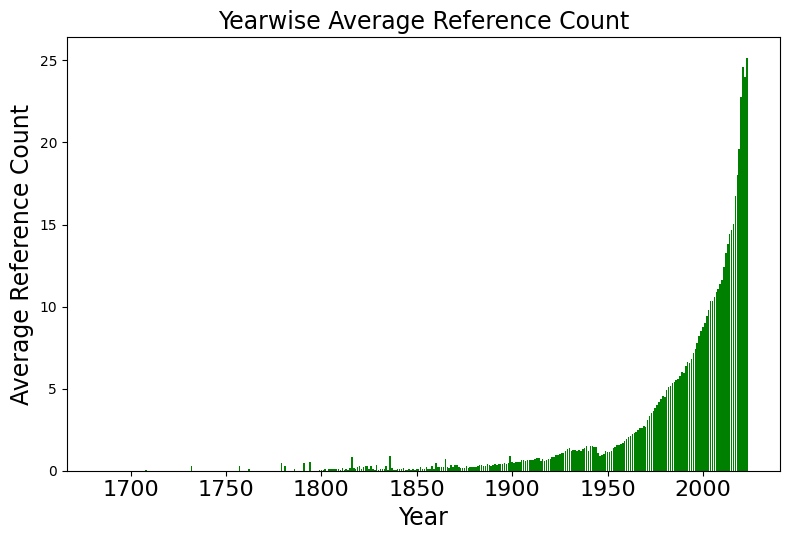}
\caption{
    The 
    year-wise average of the number of references per paper, also with a sharply increasing trend. }
        \label{fig:ref_count} 
    \end{figure}
For the Semantic Scholar dataset \citep{Kinney2023TheSS,lo-etal-2020-s2orc}, we obtain the set of 206M papers
using the ``Papers'' endpoint to get the Paper Id, Title, Abstract, Year, Citation Count, Influential Citation Count \citep{ValenzuelaEscarcega2015IdentifyingMC},and the Reference Count for each paper. 
The papers come from a variety of fields such as law, computer science, and linguistics, chemistry, material science, physics, geology etc.
For the citation network with 2.4B edges, we use the
Semantic Scholar Citations API to get each edge of the citation graph in a triplet format of (fromPaper, toPaper, isInfluentialCitations).

In general, the number of publications shows an explosive increase in recent years. \cref{fig:year_distr} shows the number of papers publish the per year, which reaches on average 7.5M per year since 2010. \cref{fig:ref_count} shows the number of references each paper cites, which also increases from less than five before 1970s, to around 25 in recent years. 
Both statistics support the need of our paper, which helps distinguish the quality of scientific studies given such massive growths of papers.

\section{Additional Analyses}
\subsection{Citation Outlier Analysis}\label{appd:outlier}

\begin{figure}[t]
    \centering
    \includegraphics[width=\linewidth]{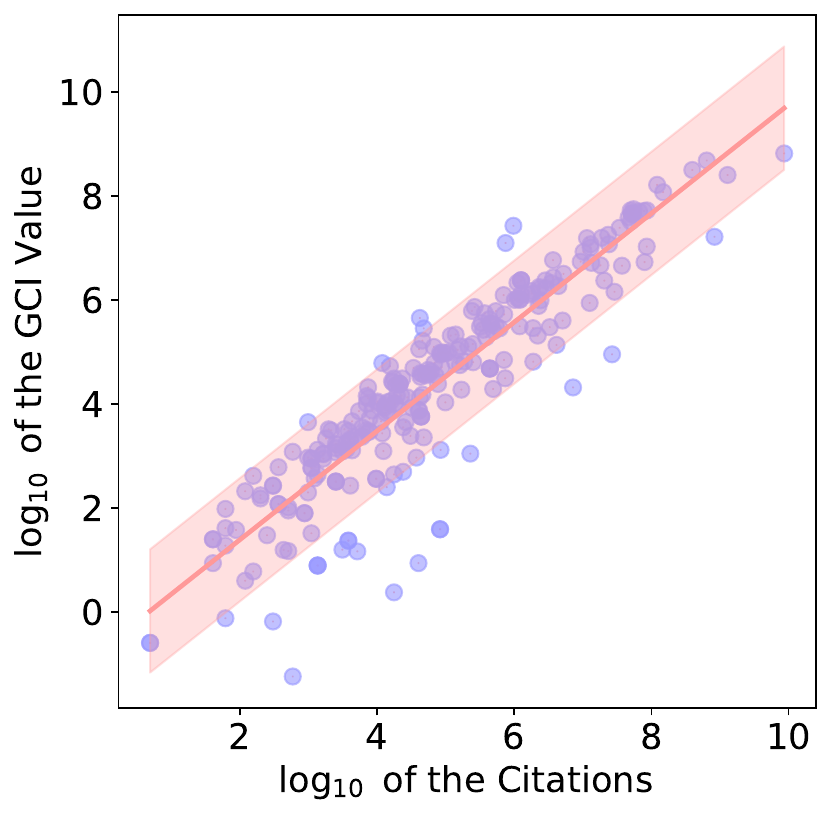}
    \caption{The scatter plot between our \ourname and citations, with the fitted function as $\log (\mathrm{TCI}) = 1.026 *  \log (\mathrm{Cit}) -0.541$, and a non-outlier band width of 0.8809.
    }
    \label{fig:correlationPlot}
\end{figure}

For the outlier detection, we first visualize the scatter plot between our \ourname and citations. Then, we fit a log-linear regression to learn the line $\log (\mathrm{TCI}) = 1.026  \log (\mathrm{Cit}) -0.541$, as shown in \cref{fig:correlationPlot}, with a root mean squared error (RMSE) of 0.6807.
After fitting the function, we use the interquartile range (IQR) method \citep{outliers},
which identify as outliers any samples that are either lower than the first quartile by over 1.5 IQR, or higher than the third quartile by more than 1.5 IQR, where IQR is the difference between the first and third quartile.

We denote as overcited papers the ones that are identified as outliers by the IQR method due to too many citations than what it should have deserved given the \ourname value. Symmetrically, we denote as undercited papers the ones that are identified as outliers by the IQR method due to too few citations than what it should have deserved given the \ourname value. And we denote the non-outlier papers as the aligned ones.

\subsection{Additional Information for the Author-Identified Paper Impact Experiment}\label{appd:zhu}
As mentioned in the main paper, the dataset is annotated by pivoting on each paper $b$, and going through each of its references $a$ to label whether $a$ has a significant influence on $b$ or not. We show an example of paper $b$ and all its 31 references in \cref{tab:zhu_example}.
We calculate the accuracy of each metric with the spirit that each non-significant paper's impact value should be lower than a significant paper's. Specifically, we go through the score of each non-significant paper, and count its accuracy as 100\% if it is lower than all the significant papers', or the more general form $n_{\mathrm{lower}} / |\mathrm{Sig}|$ of conformity, where $n_{\mathrm{lower}}$ is the number of significant papers which it is lower than, and $|\mathrm{Sig}|$ is the total number of significant papers. Then we report the overall accuracy for each score by averaging the accuracy numbers on each non-significant paper.
To illustrate the idea better, we show the calculated accuracy numbers for all three metrics on our example batch in \cref{tab:zhu_example}.

\begin{table*}[t]
    \centering \small
    \begin{tabular}{p{8.5cm}lcccc}
\toprule

\textit{References of the Paper ``Sorting improves word-aligned bitmap indexes''}                                                                                              & Label & \multicolumn{1}{l}{PCI} & \multicolumn{1}{l}{Citations} & SSHI \\ \midrule
- A Quantitative   Analysis and Performance Study for Similarity-Search Methods in   High-Dimensional Spaces & 0                           & \green{3.519}                & 1777                          & 156                        \\
- Optimizing bitmap indices with efficient   compression                                                     & 0                           & \green{3.519}                & 375                           & 40                         \\
- Data Warehouses And Olap: Concepts,   Architectures And Solutions                                          & 0                           & \green{3.526}                & \blackgreen{187}                           & 11                         \\
- Histogram-aware sorting for enhanced   word-aligned compression in bitmap indexes                          & 0                           & \green{3.543}               & \green{17}                            & \green{1}                          \\
- CubiST++: Evaluating Ad-Hoc CUBE Queries   Using Statistics Trees                                          & 0                           & \green{3.543}                & \green{5}                             & \green{1}                          \\
- Improving Performance of Sparse   Matrix-Vector Multiplication                                             & 0                           & \green{3.543}                & \blackgreen{114}                           & 11                         \\
- Binary Gray Codes with Long Bit Runs                                                                       & 0                           & \green{3.543}                & \green{53}                            & \green{4}                          \\
- Analysis of Basic Data Reordering   Techniques                                                             & 0                           & \green{3.543}                & \green{16}                            & \green{1}                          \\
- Tree Based Indexes Versus Bitmap Indexes:   A Performance Study                                            & 0                           & \green{3.543}                & \green{24}                            & \green{0}                          \\
- Secondary indexing in one dimension:   beyond b-trees and bitmap indexes                                   & 0                           & \green{3.543}                & \green{10}                            & \green{1}                          \\
- A comparison of five probabilistic   view-size estimation techniques in OLAP                               & 0                           & \green{3.543}                & \green{24}                            & \green{1}                          \\
- Compression techniques for fast external   sorting                                                         & 0                           & \green{3.543}               & \green{16}                            & \green{0}                          \\
- A Note on Graph Coloring Extensions and   List-Colorings                                                   & 0                           & \green{3.543}                & \green{33}                            & \green{1}                          \\
- Using Multiset Discrimination to Solve   Language Processing Problems Without Hashing                      & 0                           & \green{3.543}                & \green{52}                            & \green{2}                          \\
- Monotone Gray Codes and the Middle Levels   Problem                                                        & 0                           & \green{3.543}                & \blackgreen{80}                            & \blackgreen{5}                          \\
- The Art in Computer Programming                                                                            & 0                           & \green{3.543}                & 9242                          & 678                        \\
- An Efficient Multi-Component Indexing   Embedded Bitmap Compression for Data Reorganization                & 0                           & \green{3.543}                & \green{8}                             & \green{2}                          \\
- The LitOLAP Project: Data Warehousing   with Literature                                                    & 0                           & \green{3.543}                & \green{8}                             & \green{0}                          \\
- Multi-resolution bitmap indexes for   scientific data                                                      & 0                           & \green{3.583}                & \blackgreen{96}                            & \green{3}                          \\
- Notes on design and implementation of   compressed bit vectors                                             & 0                           & \green{3.583}                & \blackgreen{81}                            & 12                         \\
- Compressing Large Boolean Matrices using   Reordering Techniques                                           & 0                           & \green{3.595}                & \blackgreen{88}                            & \blackgreen{7}                          \\
- \textbf{Compressing bitmap indices by data   reorganization}                                                        & \textbf{1}                           & 3.595                & 53                            & 4                          \\
- Model 204 Architecture and Performance                                                                     & 0                           & \blackgreen{3.635}                & 238                           & \blackgreen{10}                         \\
- \textbf{On the performance of bitmap indices for   high cardinality attributes }                                    & \textbf{1}                           & 3.654                 & 196                           & 10                         \\
- A performance comparison of bitmap   indexes                                                               & 0                           & {3.655}                & \blackgreen{86}                            & \blackgreen{9}                          \\
- Minimizing I/O Costs of Multi-Dimensional   Queries with Bitmap Indices                                    & 0                           & {3.692}                & \green{16}                            & \green{0}                          \\
- Evaluation Strategies for Bitmap Indices   with Binning                                                    & 0                           & {3.692}                & \blackgreen{69}                            & \green{3}                          \\
- C-Store: A Column-oriented DBMS                                                                            & 0                           & {3.710}                & 1241                          & 111                        \\
- Byte-aligned bitmap compression                                                                            & 0                           & {3.793}                & 209                           & 48                         \\
- Bit Transposed Files                                                                                       & 0                           & {3.837}                & \blackgreen{84}                            & 10                         \\
- Space efficient bitmap indexing                                                                            & 0                           & {4.011}                & \blackgreen{96}                            & 16  
\\ \bottomrule
\end{tabular}
\caption{All the reference papers for a given study ``Sorting improves word-aligned bitmap indexes.'' 
Among all its 31 references, we \textbf{boldface} the reference papers that are annotated to be significant influencers.
For the three metrics, PCI, citations, and SSHI, we report their impact scores for each reference paper on the given study, where
we mark a score \green{in green} when it conforms to the rule that a non-significant paper's value should be lower than that of a significant paper, and mark a score \blackgreen{in dark green} if it conforms to the rule to have a lower score than one of the significant paper, but violates the rule, i.e., having a higher score than the other significant paper. In this example, our PCI metric has an accuracy score of {79.3}\%, which is higher than both citations ({68.1}\%), and SSHI ({65.0}\%).}
    \label{tab:zhu_example}
\end{table*}

\subsection{Step Curve for PCI Values Given a Fixed Paper $b$}
\begin{figure}[ht]
    \centering\includegraphics[width=\linewidth]{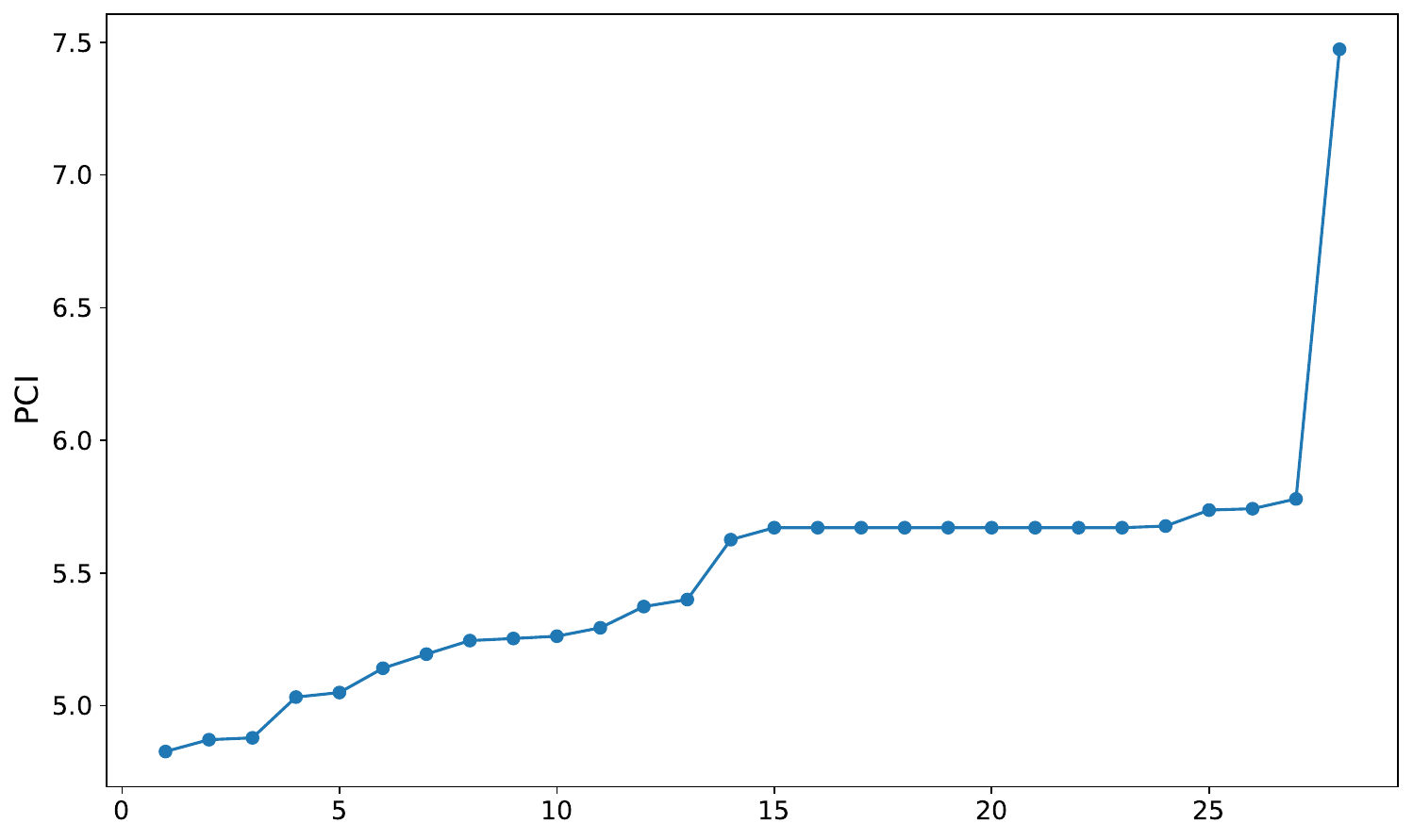}
        \caption{
        We take an example paper $b$, Sentence BERT \citep{Reimers2019SentenceBERTSE}, and plot its PCI values with all its reference paper $a$'s. We can see clearly that there is a plateau in the curve, showing a step function-like nature.
        }
        \label{fig:PCIDistributionCurve}
\end{figure}

Apart from the long-tailed curve shape of TCI in \cref{sec:curve}, we also look into the pairwise paper impacts by PCI.
If we fix the paper $b$, we can see that $\mathrm{PCI}(\cdot, b)$ often has a step curve shape in \cref{fig:PCIDistributionCurve}. The reason behind it lies in the nature of PCI, which is calculated based  on the top K papers that are similar in content with paper $b$, but do not cite paper $a$. When we go through different references, e.g., from $a_1$ to $a_2$ of the same paper $b$, the semantically matched top K papers could still be largely the same pool, and only change when some papers in the pool need to be swapped when releasing the constraint to be that they can cite $a_1$, and adding the constraint that they cannot cite $a_2$.

\end{document}